\documentclass[11pt]{article}

\usepackage[preprint]{acl}

\usepackage{times}
\usepackage{latexsym}
\usepackage[T1]{fontenc}
\usepackage[utf8]{inputenc}
\usepackage{microtype}
\usepackage{inconsolata}
\usepackage{graphicx}

\usepackage{amsmath}
\usepackage{amssymb}
\usepackage{bm}
\usepackage{booktabs}
\usepackage{multirow}
\usepackage{array}
\usepackage{subcaption}
\usepackage{algorithm}
\usepackage[noend]{algpseudocode}

\usepackage{pifont}
\newcommand{\xmark}{\text{\ding{55}}}

\usepackage[most]{tcolorbox}
\definecolor{sgopdBoxBg}{RGB}{253,243,225}
\definecolor{sgopdBoxFr}{RGB}{225,150,90}
\newtcolorbox{limitationbox}{
  enhanced, breakable,
  colback=sgopdBoxBg, colframe=sgopdBoxFr,
  boxrule=0.7pt, arc=3pt,
  left=8pt, right=8pt, top=4pt, bottom=4pt,
  before skip=4pt, after skip=4pt,
}

\definecolor{kfBoxBg}{RGB}{245,245,245}
\definecolor{kfBoxBar}{RGB}{55,55,55}
\newtcolorbox{keyfindingbox}[1]{%
  enhanced, breakable, sharp corners=downhill,
  colback=kfBoxBg, colframe=kfBoxBar,
  boxrule=0.4pt, arc=3pt,
  left=8pt, right=8pt, top=4pt, bottom=4pt,
  before skip=4pt, after skip=4pt,
  fonttitle=\bfseries\color{white},
  coltitle=white, colbacktitle=kfBoxBar,
  attach boxed title to top left={xshift=0pt, yshift=0pt},
  boxed title style={sharp corners=downhill, colback=kfBoxBar, boxrule=0pt},
  title={#1},
}

\title{SG-OPD: Sign-Gated On-Policy Distillation via \\
       Sign-Consistency Gating and Phased Teacher Sampling}

\author{
  Haoran Xu$^{1}$\thanks{Equal contribution.} \quad Hongyu Wang$^{2}$\footnotemark[1] \quad Yifei Gao$^{3}$\footnotemark[1] \quad Jiaze Li$^{1}$\thanks{Corresponding author.} \\ \textbf{Xiaofeng Zhang}$^{4}$ \quad \textbf{Xiaosong Yuan}$^{5}$ \\
  $^{1}$Zhejiang University 
  $^{2}$Hunan University
  $^{3}$Tianjin University\\
  $^{4}$Shanghai Jiao Tong University
  $^{5}$Jilin University\\
  \small{
    \textbf{Correspondence:} \href{mailto:xhr964691257@163.com}{xhr964691257@163.com}
  }
}

\begin{document}
\maketitle

\begin{abstract}
On-policy distillation (OPD) trains a student on its own trajectories with dense per-token supervision from a stronger teacher, and often outperforms off-policy distillation and standard reinforcement learning. However, we find that its effectiveness implicitly relies on two assumptions that frequently break in practice: trajectory-level alignment between the student and the teacher, and uniform token-level reliability of the teacher's preferences. We therefore propose \textbf{Sign-Gated On-Policy Distillation} (\textbf{SG-OPD}), which uses a binary verifier as a trust signal for the teacher at two complementary granularities: \emph{phased teacher sampling} mixes in verifier-endorsed teacher rollouts at cold-start, and a \emph{sign-consistency gate} extrapolates the distillation update on tokens where the teacher agrees with the verifier-correct direction and interpolates it where it disagrees. Experiments on competition-level mathematical reasoning benchmarks show that SG-OPD consistently outperforms standard OPD, with average gains of $1.98$ and $7.50$ at the per-sample and per-question levels, respectively.
\end{abstract}


\section{Introduction}
\label{sec:intro}

The strong reasoning capabilities of large language
models~\citep{guo2025deepseekr1, yang2025qwen3} come at steep
computational cost, motivating
distillation~\citep{hinton2015distilling} to compress them into
smaller students. Off-policy
distillation~\citep{taori2023alpaca} trains on
teacher-generated trajectories but
suffers from exposure bias~\citep{bengio2015scheduled} at
inference time. On-policy distillation
(OPD)~\citep{agarwal2024gkd, lu2025opd} resolves this mismatch
by sampling from the student and minimising
a reverse KL to the teacher, yielding dense per-token
supervision on the student's own distribution.

However, we observe that the effectiveness of OPD implicitly
relies on two assumptions that frequently break in practice,
which we attribute to the following two structural
limitations:

\begin{limitationbox}
\begin{itemize}\itemsep -0.05em\leftskip-1em
\item \emph{Trajectory-level alignment is fragile.} In
strong-to-weak settings, a weak student's cold-start rollouts lie
far from the teacher's trajectory support, so the reverse-KL
signal there is dominated by noise rather than informative
supervision.
\item \emph{Token-level teacher reliability is not uniform.}
Even for rollouts verified as correct, the teacher's per-token
preferences may contradict the verifier-correct direction,
causing the reverse-KL gradient to penalize tokens that support
a valid trajectory.
\end{itemize}
\end{limitationbox}

\begin{figure*}[t]
\centering
\includegraphics[width=0.9 \textwidth]{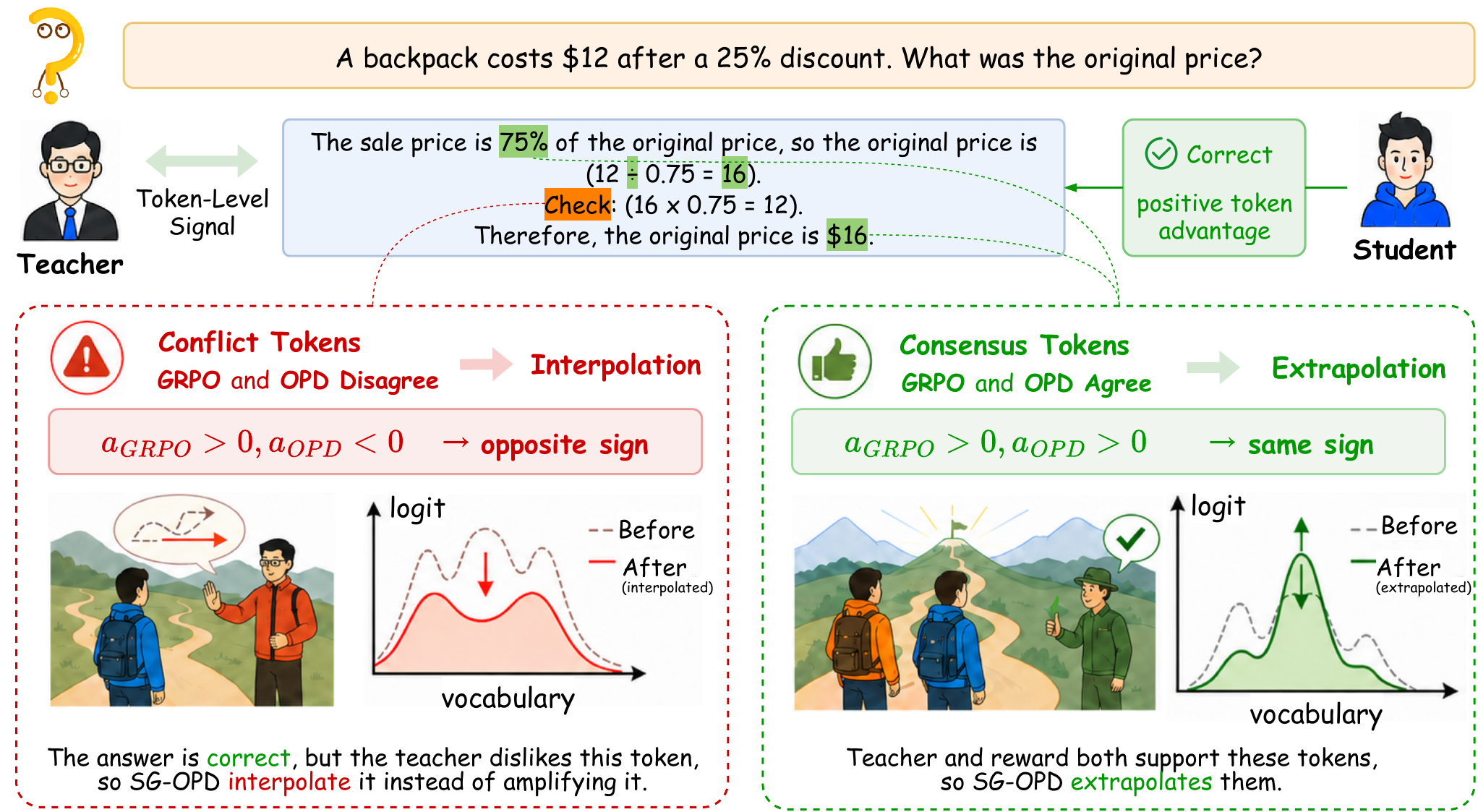}
\caption{Illustration of token-level sign-consistency gating on a
correct math rollout. Since the final answer is verified as correct,
the outcome-level GRPO advantage $a_1$ is positive for the trajectory.
However, the OPD reverse-KL advantage $a_2$ can still vary by token:
consensus tokens such as ``75\%'', ``$\div$'', and ``16'' have
$a_1 a_2>0$ and are extrapolated, while a redundant verification token
such as ``Check:'' can receive $a_2<0$ because the teacher assigns it
lower probability, so SG-OPD routes it through interpolation.}
\label{fig:conflict_schematic}
\end{figure*}

We propose
\textbf{Sign-Gated On-Policy Distillation (SG-OPD)}, which
treats a binary verifier as a trust signal for the teacher at two complementary granularities,
as illustrated in Figure~\ref{fig:conflict_schematic}.
Specifically,

\textbf{Sample level: Phased Teacher Sampling.} To bridge the
trajectory-level mismatch at cold-start, we adopt the optimization strategy inspired by mixed policy optimization\citep{lin2025chord} instead of the conventional SFT-then-by-RL paradigm\citep{xiao2026recipe}. We use the annealed schedule
mixes in verifier-endorsed teacher rollouts early in training
and decays to fully on-policy student rollouts later, so that
the student's distillation targets are drawn from trajectories
the verifier deems correct precisely when its own rollouts are
not yet aligned with the teacher.

\textbf{Token level: Sign-Consistency Gate.} To enforce
per-token reliability, we combine the verifier outcome with
the reverse-KL advantage to label each token as either
\emph{consensus}, when the teacher agrees with a
verifier-correct direction, or \emph{conflict}, when the
teacher would have moved the student away from a
verifier-correct trajectory. Consensus tokens are extrapolated
to amplify the trustworthy distillation signal, while conflict
tokens are interpolated to mute them.

Together, the two mechanisms let the student inherit teacher
supervision where the teacher is reliable, and back
off where it is not.

Under a strong-to-weak setup on competition-level math reasoning
benchmarks, SG-OPD delivers stronger and more robust performance
than existing baselines, remaining stable where uniform
extrapolation collapses.
Our main contributions are:
\begin{itemize}\itemsep -0.05em
\item We identify two implicit assumptions of on-policy
distillation that frequently break in practice: a
trajectory-level alignment assumption, where student rollouts
and teacher trajectories are insufficiently aligned at
cold-start, and a token-level reliability assumption, where
the teacher's per-token preferences contradict verifier-correct
directions even on rollouts judged correct by the verifier.
\item We propose \textbf{SG-OPD}, which uses a binary verifier
purely as a trust signal for the teacher, combining phased
teacher sampling at the trajectory level with
sign-consistency-gated extrapolation/interpolation at the
token level.
\item Experiments on competition-level mathematical reasoning
benchmarks show that SG-OPD consistently improves over
existing on-policy distillation baselines.
\end{itemize}

\section{Related Work}
\label{sec:related}

\paragraph{On-policy distillation.}
Classical KD~\citep{hinton2015distilling} fits the student to a
frozen teacher, typically via SFT on teacher
responses~\citep{taori2023alpaca} or sequence-level variants like
SeqKD~\citep{kim2016sequence}. OPD~\citep{lu2025opd} instead samples from the student
and minimizes the reverse KL, providing dense
per-token feedback at the cost of student--teacher mismatch. Video-OPD ~\citep{li2026videoopdefficientposttrainingmultimodal} tackles the challenge of cross-modal misalignment.
G-OPD~\citep{yang2026learning} recasts OPD as a
KL-regularized RL problem with a single global extrapolation factor
$\lambda$, and AOPD~\citep{jia2026asymmetriconpolicydistillationbridging} switches negative-advantage
tokens from policy gradient to truncated forward-KL. These methods either share a global $\lambda$ or
condition only on a distillation-side signal. Several works~\citep{li2026rethinkingonpolicydistillationlarge} have sought to elucidate the factors driving the performance gains of the OPD approach.


\paragraph{Mixed Policy optimization.}
GRPO~\citep{shao2024grpo} and its
successors~\citep{guo2025deepseekr1, liu2025dapo} optimize a binary
verifiable reward with group-normalized advantages, in the spirit
of classical trust regions~\citep{schulman2017proximal}. Recent
work has been exploring mixed policy optimization by integrating SFT and GRPO. For example, ExPO~\citep{zheng2025expo} extrapolates model weights post-hoc and
CHORD~\citep{lin2025chord} re-weights an SFT loss with a fixed
prior, while teacher-trajectory mixing
\citep{hejna2024hybrid} bridges cold-start with
off-policy expert data. DFT\citep{wu2026generalizationsftreinforcementlearning} is analogous. 



\section{Preliminaries and Failure Modes of OPD}
\label{sec:prelim}

Let $\pi_\theta$ be the student, $\pi^*$ the frozen teacher, and
$\pi_{\mathrm{ref}}$ the reference policy initialized from the
student. A prompt $x\!\sim\!\mathcal{D}$ generates a student
trajectory $y\!=\!(y_1,\ldots,y_T)$ with verifiable outcome reward
$r(x,y)\!\in\!\{0,1\}$. 

\paragraph{Verifier signal (sample level).}
We define an outcome-level
\textbf{verifier signal}
$a_1(t)\!:=\!(r(x,y)\!-\!\mu_x)/(\sigma_x\!+\!\epsilon)$
computed in the GRPO style, normalized across $G$ rollouts of
$x$ with mean and std $\mu_x,\sigma_x$. With
binary $r$, $a_1$ is constant per trajectory; \emph{we use
it only as a trust signal for the teacher}, not as an
optimization target.

\paragraph{OPD's per-token signal.}
The mechanism we want to stabilize is the OPD policy gradient. At each token $y_t$, OPD provides the
\textbf{reverse-KL advantage}
$a_2(t)\!:=\!\log\pi_\theta(y_t)-\log\pi^*(y_t)$,
derived from the on-policy reverse-KL objective under a per-token discount of
$0$~\citep{lu2025opd}. The OPD policy gradient
is then the dense per-token form
\begin{equation}
\nabla_\theta \mathcal{J}_{\mathrm{OPD}}
= \mathbb{E}\!\left[\sum_{t=1}^{T}
  a_2(t)\,\nabla_\theta \log \pi_\theta(y_t)\right].
\label{eq:opd-grad}
\end{equation}

\paragraph{G-OPD extrapolation.}
\citet{yang2026learning} reinterpret this objective as
KL-regularized RL with extrapolation factor $\lambda\!\ge\!1$
and obtain a \textbf{G-OPD} advantage
$A_t^{\mathrm{G\text{-}OPD}}(\lambda)$ generalising $a_2$;
$\lambda\!=\!1$ recovers OPD, $\lambda\!>\!1$ extrapolates
beyond the teacher (\textbf{ExOPD}), and larger $\lambda$
degrades training in our setting. The full reverse-KL
objective and the explicit G-OPD form are deferred to
Appendix~\ref{appx:derivation}.

\paragraph{Failure mode 1: trajectory-level alignment is
fragile.} OPD assumes that the student's rollouts and the
teacher's trajectories are sufficiently aligned. In
strong-to-weak settings, a weak student often produces
trajectories that the teacher itself would find unlikely, so
the reverse-KL signal at cold-start is unreliable and the
early distillation update is dominated by noise rather than
informative supervision. Pushing $\lambda$ beyond a moderate
value amplifies this noise: the
``untrainable'' regime of \citet{yang2026learning} is observed precisely
when the early student distribution sits far from the
teacher's.

\paragraph{Failure mode 2: token-level teacher reliability is
not uniform.} OPD also assumes that the teacher is uniformly
trustworthy along every student rollout. We find that this
assumption breaks even on rollouts that the verifier judges
correct: while $a_1(t)\!>\!0$ for the entire trajectory, $a_2$
can still flip sign per token, indicating that the teacher
would have suppressed a token that lies on a verifiably correct
path. We refer to tokens with $a_1(t)\,a_2(t)\!\le\!0$ as
\emph{conflict} tokens; on these tokens, blindly amplifying the
reverse-KL gradient drives the student away from a verified
solution. Fig.~\ref{fig:conflict_schematic} visualises this
per-token pattern.

\paragraph{Empirical signature.}
Two observations make these concrete.
\textbf{(i)} The reverse-KL signal at cold-start is dominated
by trajectory-level mismatch, and uniformly increasing
$\lambda$ does not recover performance
(Tab.~\ref{tab:main}).
\textbf{(ii)} A non-trivial fraction of tokens remain in the
conflict regime $a_1 a_2 \!\le\! 0$ throughout training rather
than only at cold-start
(Fig.~\ref{fig:sign-agree}, Appendix~\ref{appx:case-study}).
Sec.~\ref{sec:method} introduces a two-granularity framework
that uses the verifier signal $a_1$ as a trust signal for the
teacher to address both failure modes.

\section{Method: SG-OPD}
\label{sec:method}

SG-OPD couples verifiable-reward RL with OPD, using a binary
verifier as a trust signal at two levels:
(i) a \emph{sample-level} teacher anchor, \textbf{phased
teacher sampling (PTS)}, adding an auxiliary loss on verified
teacher rollouts; and (ii) a \emph{token-level} sign-consistency
mechanism, routing each token by whether the
verifier-induced advantage and the OPD advantage
agree in sign. Algorithm~\ref{alg:sg-opd} summarizes the full
training step.

\paragraph{Notation recap.}
We collect the symbols below (all introduced in \S\ref{sec:prelim}).
\begin{center}
\resizebox{\columnwidth}{!}{%
\begin{tabular}{@{}ll@{}}
\toprule
Symbol & Meaning \\
\midrule
$\pi_\theta,\pi^*,\pi_{\mathrm{ref}}$ & student / frozen teacher / reference policy \\
$x,\;y\!=(y_1,\ldots,y_T)$ & prompt and student rollout \\
$r(x,y)\!\in\!\{0,1\}$ & verifiable outcome reward \\
$a_1(t)$ & GRPO outcome advantage at token $t$ \\
$a_2(t)$ & OPD reverse-KL advantage at token $t$ \\
$g_t$ & token-level sign-consistency gate \\
$\phi_t$ & detached stability weight \\
$\alpha(t)$ & phased teacher-anchor schedule \\
$\mathcal{L}_{\mathrm{TLT}}$ & token-level sign-consistency-gated loss \\
$\mathcal{L}_{\mathrm{SLT}}$ & sample-level teacher-anchor loss \\
\bottomrule
\end{tabular}%
}
\end{center}
Hyperparameters fall into five groups: \emph{extrapolation strength}
($\lambda_{\mathrm{high}}$, $\lambda_{\mathrm{base}}$),
\emph{conflict fallback} ($\beta$ and the fallback mode),
\emph{teacher sampling ratio} ($\rho$), \emph{phased schedule}
($P_1,P_2,\alpha_0,\alpha_{\mathrm{end}}$), and
\emph{stability clipping} ($\tau$). Default values are listed in
Appendix~\ref{appx:hyperparams}.

\subsection{Sample-Level Teacher Anchor: Phased Teacher Sampling (PTS)}
\label{sec:method-slt}

\paragraph{Motivation.}
At cold-start the student attains low accuracy, so most on-policy rollouts are incorrect.then dominated by trajectories the teacher itself would not support,
and the student can drift onto a low-reward manifold where the
verifier supplies little corrective gradient. PTS addresses this
sample-level failure mode by injecting a small number of verified
teacher rollouts early in training and then annealing this anchor
away.

\begin{figure*}[t]
\centering
\includegraphics[width=\textwidth]{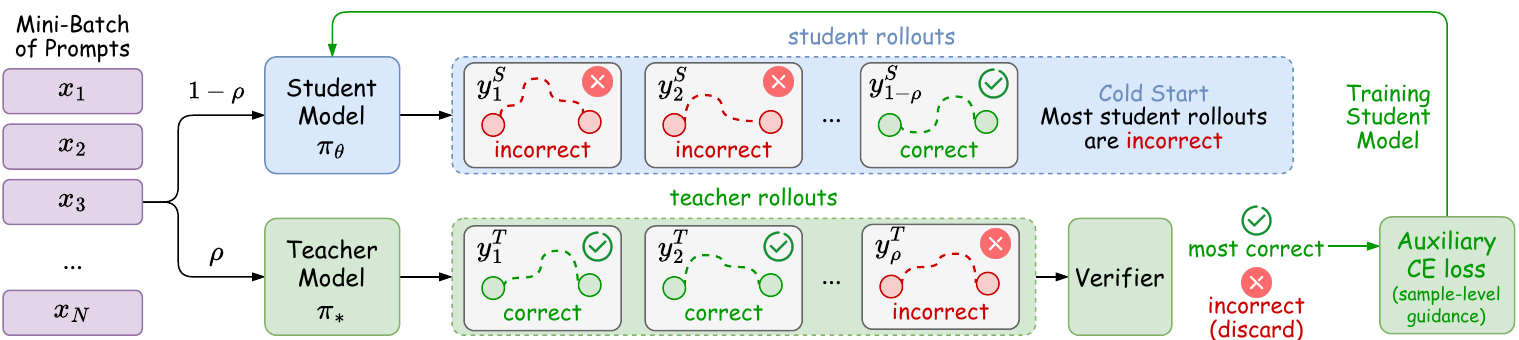}
\caption{Sample-level Phased Teacher Sampling (PTS). A mini-batch is
split into student on-policy rollouts and a small fraction of teacher
rollouts. Verified teacher trajectories are retained
and used as an auxiliary CE anchor, while incorrect teacher
trajectories are discarded. The teacher-guidance weight is annealed
from warm-up to zero, so the asymptotic training distribution remains
on-policy.}
\label{fig:method-pts-overview}
\end{figure*}

\paragraph{Verified teacher rollouts.}
For each mini-batch $\mathcal{B}$, we reserve a fraction $\rho$ of
prompts for teacher sampling, yielding $\mathcal{B}_T$. The teacher
generates $y^T\!\sim\!\pi^*(\cdot|x)$ on this subset, and only
trajectories verified as correct are retained. The resulting
sample-level teacher-anchor loss is
\begin{equation}
\mathcal{L}_{\mathrm{SLT}}
= \sum_{\substack{(x,y^T)\in \mathcal{B}_T \\ r(x,y^T)=1}}
  \sum_{t}
  \mathcal{L}_{\mathrm{CE}}\!\bigl(\pi_\theta, y^T_t\bigr).
\label{eq:slt}
\end{equation}
Incorrect teacher rollouts are discarded, so the anchor is defined
by verifier agreement rather than teacher likelihood alone.

\paragraph{Phased annealing.}
The teacher anchor is useful during cold-start but should not define
the asymptotic training distribution. We therefore weight it by a
three-phase cosine schedule,
\begin{equation}
\resizebox{\columnwidth}{!}{$\displaystyle
\alpha(t) =
\begin{cases}
\alpha_0, & t \le P_1, \\[3pt]
\alpha_{\mathrm{end}} + \tfrac{\alpha_0 - \alpha_{\mathrm{end}}}{2}
\!\left(1 + \cos\!\left(\tfrac{\pi(t-P_1)}{P_2-P_1}\right)\right), & P_1 < t \le P_2, \\[3pt]
0, & t > P_2,
\end{cases}
$}
\label{eq:phase-schedule}
\end{equation}
where $\alpha_0$ is the cold-start weight, $\alpha_{\mathrm{end}}$
is the value at the end of the transition window, and $P_1,P_2$ are
the phase boundaries. Once $\alpha(t)\!=\!0$, the auxiliary anchor is
removed and training becomes fully on-policy.

\subsection{Token-Level Sign-Consistency and Stability Weighting}
\label{sec:method-tlt}

\paragraph{Motivation.}
PTS controls \emph{which trajectories} the student visits. It does
not resolve token-level sign conflict, which arises within a fixed
student rollout: on some tokens, the verifier-induced GRPO advantage
$a_1(t)$ and the teacher-induced OPD advantage $a_2(t)$ point in
opposite directions. A uniform linear combination then amplifies an
update direction opposed by one of the two signals. SG-OPD therefore
routes tokens by sign agreement before forming the policy-gradient
advantage.

\begin{figure}[t]
\centering
\includegraphics[width=\columnwidth]{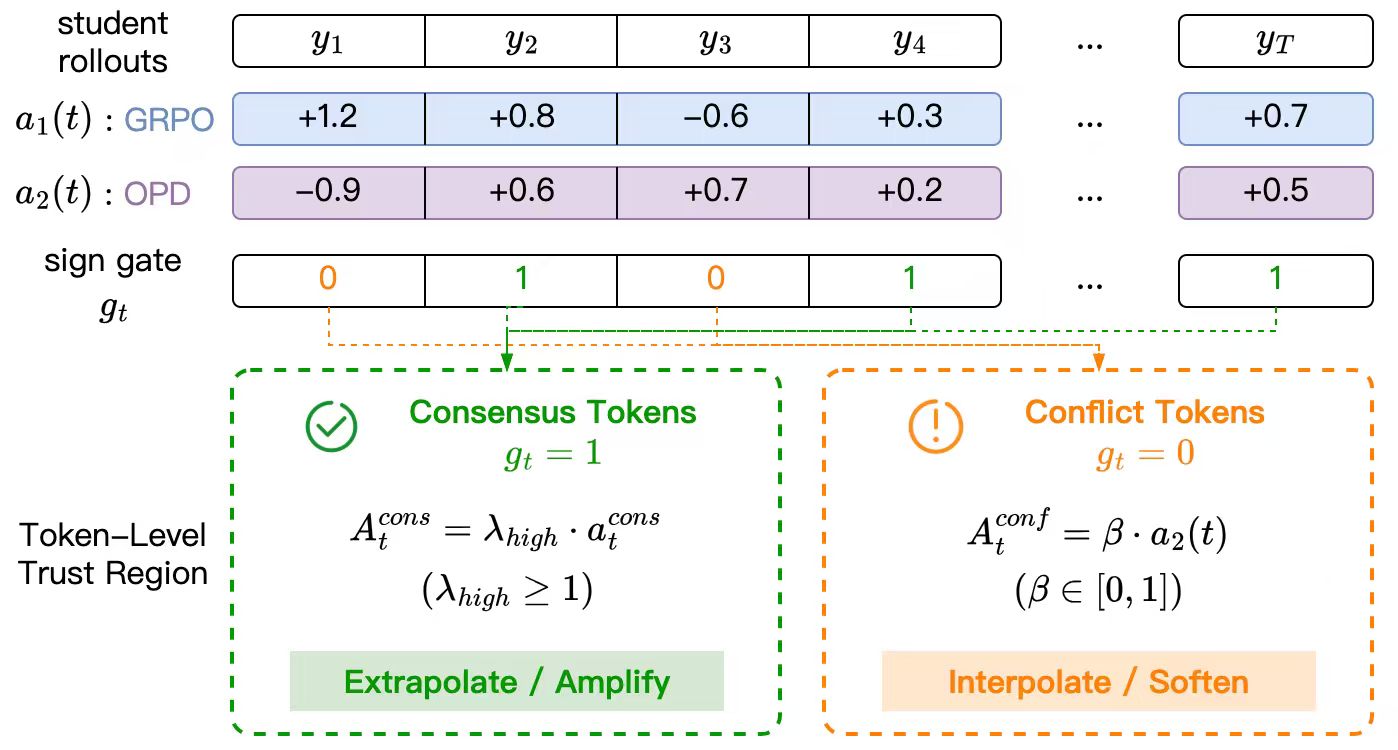}
\caption{Overview of token-level sign-consistency gating. GRPO and OPD token advantages are routed by sign agreement:
consensus tokens are extrapolated, while conflict tokens are softened by interpolation.}\label{fig:method-signgate-overview}
\end{figure}

\paragraph{Sign-consistency gate.}
We encode whether the two token-level signals agree with
\begin{equation}
g_t \;:=\; \mathbf{1}\!\left[\, a_1(t) \cdot a_2(t) > 0 \,\right]
\;\in\;\{0, 1\}.
\label{eq:sign-gate}
\end{equation}
Here $g_t\!=\!1$ marks \emph{consensus} tokens, where the verifier
and teacher push the sampled token in the same direction, while
$g_t\!=\!0$ marks \emph{conflict} tokens.

\paragraph{Routed token advantage.}
Let
\begin{multline}
A_t^{\mathrm{G\text{-}OPD}}(\lambda)
= a_2(t) \\
+ (\lambda\!-\!1)\bigl(\log\pi_{\mathrm{ref}}(y_t)
      -\log\pi^*(y_t)\bigr),
\label{eq:gopd-adv}
\end{multline}
which recovers OPD at $\lambda\!=\!1$ and ExOPD at
$\lambda\!>\!1$. On consensus tokens we apply stronger
extrapolation,
\begin{equation}
A_t^{\mathrm{cons}}
= A_t^{\mathrm{G\text{-}OPD}}(\lambda_{\mathrm{high}}),
\qquad \lambda_{\mathrm{high}}>1,
\label{eq:cons}
\end{equation}
and on conflict tokens the default fallback is softened OPD,
\begin{equation}
A_t^{\mathrm{conf}} = \beta\,a_2(t),\qquad \beta\!\in\![0,1].
\label{eq:conf-interp}
\end{equation}
Thus $\beta\!=\!1$ recovers OPD on conflict tokens, whereas
$\beta\!=\!0$ masks them. Alternative \texttt{preserve} and
\texttt{grpo} fallbacks are evaluated in Tab.~\ref{tab:full-ablation}.
The routed advantage is
\begin{equation}
A_t^{\mathrm{SG}}
= g_t \cdot A_t^{\mathrm{cons}}
+ (1 - g_t) \cdot A_t^{\mathrm{conf}}.
\label{eq:sg-adv}
\end{equation}
When the gate is disabled, $A_t^{\mathrm{SG}}$ reduces to G-OPD,
so this strictly generalizes uniform extrapolation.

\paragraph{Stability weighting.}
Large $|a_2(t)|$ values can cause a small number of OPD outliers to
dominate the actor gradient. We therefore apply a detached clipping
weight after the sign-consistency decision,
\begin{equation}
\phi_t \;=\; \min\bigl(1,\;\tau / |a_2(t)|\bigr),
\label{eq:conf-weight}
\end{equation}
where $\tau$ is a clipping hyperparameter. In implementation,
$\phi_t$ is detached and batch-normalized to unit mean. It therefore
changes the scale of the token update but not whether a token is
classified as consensus or conflict. Ablations in
Appendix~\ref{appx:full-ablation} isolate this weighting from the
sign-consistency gate.

\subsection{Combined Objective and Algorithm}
\label{sec:method-loss}

The token-level policy-gradient loss is
\begin{equation}
\mathcal{L}_{\mathrm{TLT}}(\theta)
= -\,\mathbb{E}\!\left[\tfrac{1}{|y|}\!\sum_{t=1}^{|y|}
   \phi_t\,A_t^{\mathrm{SG}}\,\log\pi_\theta(y_t\!\mid\!c_t)\right],
\label{eq:tlt}
\end{equation}
where the standard PPO importance-ratio clip~\citep{schulman2017proximal}
replaces the $\log\pi_\theta$ factor in our implementation. The
complete objective is
\begin{equation}
\mathcal{L}_{\mathrm{SG\text{-}OPD}}(\theta)
= \mathcal{L}_{\mathrm{TLT}}(\theta)
+ \alpha(t)\,\mathcal{L}_{\mathrm{SLT}}(\theta).
\label{eq:total-loss}
\end{equation}
Equivalently, the token-level term can be written as the routed
per-token expectation
\begin{align}
&\mathcal{L}_{\mathrm{SG\text{-}OPD}}(\theta)
= -\,\mathbb{E}\!\Bigl[\tfrac{1}{|y|}\!\sum_{t=1}^{|y|}
   \phi_t\bigl(g_t\,A_t^{\mathrm{cons}}\nonumber\\
&\quad+ (1\!-\!g_t)\,A_t^{\mathrm{conf}}\bigr)
   \log\pi_\theta(y_t\!\mid\! c_t)\Bigr]\nonumber\\
&\quad+ \alpha(t)\,\mathcal{L}_{\mathrm{SLT}}(\theta).
\label{eq:total-loss-explicit}
\end{align}
The token-level gate is always active, while the sample-level
teacher anchor is phased out by $\alpha(t)$. Thus SG-OPD uses
verified teacher rollouts to stabilize cold-start but restores a
fully on-policy objective once the anchor is annealed away.

\begin{algorithm}[t]
\small
\caption{One step of SG-OPD.}
\label{alg:sg-opd}
\begin{algorithmic}[1]
\Require Student $\pi_\theta$, teacher $\pi^*$, reference
  $\pi_{\mathrm{ref}}$, current step $t$, total steps $T$,
  ratio $\rho$, phase $(P_1, P_2)$,
  $(\lambda_{\mathrm{base}}, \lambda_{\mathrm{high}}, \beta, \tau)$.
\State Sample mini-batch $\mathcal{B}\!=\!\{x_i\}$ from $\mathcal{D}$.
\State Split $\mathcal{B}$ into $\mathcal{B}_S$ and $\mathcal{B}_T$
  with $|\mathcal{B}_T|/|\mathcal{B}|\!=\!\rho$.
\State $y_S \!\gets\! \pi_\theta(\cdot|x_i)$ for
  $x_i\!\in\!\mathcal{B}_S$ \Comment{on-policy rollouts}
\State $y_T \!\gets\! \pi^*(\cdot|x_j)$ for
  $x_j\!\in\!\mathcal{B}_T$ \Comment{teacher rollouts}
\State Verify $r(x,y)\!\in\!\{0,1\}$ on all trajectories.
\For{token $y_t$ in $y_S$}
  \State $a_1(t) \!\gets\! A_t^{\mathrm{GRPO}}$,\;
         $a_2(t) \!\gets\! \log\pi_\theta(y_t)\!-\!\log\pi^*(y_t)$.
  \State $g_t \!\gets\! \mathbf{1}[a_1\!\cdot\!a_2 \!>\!0]$
         \Comment{sign-consistency gate}
  \State $A_t^{\mathrm{SG}} \!\gets\!$ Eq.~\eqref{eq:sg-adv};
         $\phi_t \!\gets\!$ Eq.~\eqref{eq:conf-weight}.
\EndFor
\State $\mathcal{L}_{\mathrm{TLT}} \!\gets\!$ PPO-clipped policy
       gradient with $A_t^{\mathrm{SG}} \cdot \phi_t$.
\State Filter $y_T$ by $r(x_j, y_T)\!=\!1$.
\State $\mathcal{L}_{\mathrm{SLT}} \!\gets\!$ Eq.~\eqref{eq:slt}.
\State $\alpha(t)\!\gets\!$ Eq.~\eqref{eq:phase-schedule}.
\State $\mathcal{L} \!\gets\! \mathcal{L}_{\mathrm{TLT}}
       + \alpha(t)\,\mathcal{L}_{\mathrm{SLT}}$.
\State Update $\theta \!\gets\! \theta - \eta\,\nabla_\theta\mathcal{L}$.
\end{algorithmic}
\end{algorithm}

The sign-consistency gate modifies the actor-update path, whereas
PTS adds the teacher-rollout and SFT-loss path. The full algorithm
is implemented in the verl framework~\citep{verl2024}; we will
release the implementation upon acceptance.

\label{sec:exp}

Our experiments are organized around four questions:
\begin{itemize}
\setlength{\itemsep}{1pt}
\setlength{\parskip}{0pt}
\item \textbf{(Q1)} Does SG-OPD improve over OPD and ExOPD on
competition-level math reasoning under an identical recipe?
(\S\ref{sec:main-results}, Tab.~\ref{tab:main})
\item \textbf{(Q2)} Do the sample-level and token-level
mechanisms each contribute on their own, and are they
complementary? (\S\ref{sec:ablation})
\item \textbf{(Q3)} Does the sign-consistency gate widen the
safe range of consensus-token extrapolation strength, recovering
performance at a $\lambda_{\mathrm{high}}$ where uniform
extrapolation collapses?
(\S\ref{sec:ablation}, Tab.~\ref{tab:full-ablation})
\item \textbf{(Q4)} Does SG-OPD raise training reward without
collapsing policy entropy, as the token-level gate predicts?
(\S\ref{sec:dynamics}, Fig.~\ref{fig:baseline-dynamics})
\end{itemize}

\section{Experiments}
\subsection{Setup}
\label{sec:exp-setup}

\paragraph{Models.}
The student $\pi_\theta$ is Qwen3-1.7B-Non-Thinking, and the teacher
$\pi^*$ is the step-500 Qwen3-4B-Non-Thinking-RL-Math checkpoint. The
reference $\pi_{\mathrm{ref}}$ is the student's initial state. All
three share the same tokenizer, following the strong-to-weak setting
in Table~3.

\paragraph{Training data.}
Training prompts come from DeepMath-103K~\citep{deepmath2025} filtered
to difficulty level $\!\ge\!6$, resulting in $57\,$K problems. For each
prompt we sample $G\!=\!8$ rollouts at temperature $1.0$.

\paragraph{Evaluation.}
We evaluate on four competition-level math reasoning benchmarks:
AIME24, AIME25, HMMT25-Feb, and HMMT25-Nov. We report avg@$32$ and
pass@$32$ accuracy with sampling temperature $\mathcal{T}\!=\!1.0$,
top-$p\!=\!1.0$, and a generation budget of $16{,}384$ tokens; AVG
always denotes the arithmetic mean over the benchmarks. This protocol matches \citet{yang2026learning}'s Table~3.

\paragraph{Training and baselines.}
We train for $100$ optimizer steps with the GRPO advantage
estimator and no learned critic. SG-OPD default hyperparameters are
selected via the ablations in \S\ref{sec:ablation}; OPD and ExOPD
baselines are trained under the same recipe. All schedule
comparisons use the same optimizer and teacher-rollout budgets.
Full optimization details, hyperparameters, and run-to-run variance
are in Appendix~\ref{appx:hyperparams} and~\ref{appx:reproducibility}.

\subsection{Main Results (Q1)}
\label{sec:main-results}

\begin{table*}[t]
\footnotesize
\centering
\setlength{\tabcolsep}{2pt}
\renewcommand{\arraystretch}{1.05}
\resizebox{\textwidth}{!}{%
\begin{tabular}{@{}l|cccc|c|cccc|c@{}}
\toprule
\multirow{2}{*}{\textbf{Method}}
& \multicolumn{5}{c|}{\textbf{avg@$32$ (\%)}}
& \multicolumn{5}{c@{}}{\textbf{pass@$32$ (\%)}} \\
\cmidrule(lr){2-6} \cmidrule(lr){7-11}
& AIME24 & AIME25 & HMMT25-Feb & HMMT25-Nov & \textbf{AVG}
& AIME24 & AIME25 & HMMT25-Feb & HMMT25-Nov & \textbf{AVG} \\
\midrule
SFT~\citep{taori2023alpaca}              & 30.93 & 25.62 & 16.04 & 18.33 & 22.73 & 48.33 & 33.67 & 24.67 & 31.41 & 34.52 \\
SeqKD~\citep{kim2016sequence}            & 31.88 & 26.46 & 16.46 & 18.96 & 23.44 & 50.00 & 36.67 & 26.67 & 33.33 & 36.67 \\
GKD~\citep{agarwal2024gkd}               & 37.71 & 32.19 & 17.40 & 18.96 & 26.57 & 66.67 & 50.00 & 36.67 & \underline{46.67} & 50.00 \\
OPD~\citep{lu2025opd}                    & 38.96 & \underline{33.44} & \underline{18.02} & \underline{19.79} & 27.55 & \underline{70.00} & 50.00 & 40.00 & \underline{46.67} & 51.67 \\
ExOPD~\citep{yang2026learning}           & \underline{40.83} & 33.12 & \textbf{20.21} & 17.81 & \underline{27.99} & \underline{70.00} & \underline{53.33} & \textbf{50.00} & 43.33 & \underline{54.17} \\
\midrule
\textbf{SG-OPD (Ours)}                   & \textbf{41.35} & \textbf{38.44} & \underline{18.02} & \textbf{20.31} & \textbf{29.53}
                                         & \textbf{76.67} & \textbf{66.67} & \underline{43.33} & \textbf{50.00} & \textbf{59.17} \\
\textit{$\Delta$ vs OPD}                 & \textit{+2.39} & \textit{+5.00} & \textit{$\pm$0.00} & \textit{+0.52} & \textit{+1.98}
                                         & \textit{+6.67} & \textit{+16.67} & \textit{+3.33} & \textit{+3.33} & \textit{+7.50} \\
\bottomrule
\end{tabular}%
}
\caption{Main results across four competition-level math reasoning
benchmarks. We compare \textbf{SG-OPD} against
SeqKD, GKD, OPD, and ExOPD under the strong-to-weak setting
(Qwen3-1.7B distilled from Qwen3-4B-Non-Thinking-RL-Math, step 500).
\textbf{Bold} marks the best result within each column, while
\underline{underlined} values denote the second best;
\textit{$\Delta$ vs OPD} reports the absolute improvement of
SG-OPD over the OPD baseline.}
\label{tab:main}
\end{table*}

\begin{keyfindingbox}{Key Finding 1}
A binary verifier provides a reliable token-level trust signal for
the teacher in on-policy distillation.
\end{keyfindingbox}

Table~\ref{tab:main} summarizes the main results.
\textbf{SG-OPD achieves the highest AVG under both metrics}: $29.53$ avg@$32$
($+1.98$ over OPD, $+1.54$ over ExOPD) and $59.17$ pass@$32$ ($+7.50$
over OPD, $+5.00$ over ExOPD). The gains vary across
benchmarks, and the per-benchmark breakdown reveals where
SG-OPD's two mechanisms contribute most.

\paragraph{Per-benchmark trends (avg@$32$).}
On AIME-style benchmarks, the largest gain appears on AIME25 ($+5.00$
over OPD), where the sign-conflict fraction is also highest
(\S\ref{sec:dynamics}); AIME24 shows a smaller but consistent
$+2.39$ improvement. The HMMT subsets pose greater difficulty (avg@$32$ in the
$18$--$20\%$ range) and SG-OPD matches OPD exactly on HMMT25-Feb
($18.02$) and improves slightly on HMMT25-Nov ($+0.52$).

\paragraph{Pass@$32$ unlocks the largest improvements.}
As shown in the pass@$32$ block of Table~\ref{tab:main}, when counting
trajectories that successfully solve the problem at least once, the
exploration advantage of SG-OPD becomes more pronounced: SG-OPD reaches
$76.67$ on AIME24 ($+6.67$ over OPD) and $66.67$ on AIME25 ($+16.67$). This is
consistent with the design of SG-OPD
(Sec.~\ref{sec:method}): the gate permits aggressive extrapolation
on consensus tokens, expanding the set of trajectories the student
can reach without sacrificing the verifier's outcome signal on
conflict tokens.

\subsection{Ablations}
\label{sec:ablation}

We conducted over $50$ controlled training runs varying the
token-level and sample-level hyperparameters. Four representative
slices are reported here; the full table is in Appendix~\ref{appx:full-ablation}.

\begin{keyfindingbox}{Key Finding 2}
The sign-consistency gate widens the safe range of consensus-token
extrapolation strength.
\end{keyfindingbox}

\paragraph{Token-level gate (Q3).}
Tab.~\ref{tab:full-ablation} sweeps the token-level knobs
$\lambda_{\mathrm{high}}$, fallback mode, and $\beta$.
The configuration maximizing AVG employs $\lambda_{\mathrm{high}}\!=\!1.8$ with
\texttt{interp} and $\beta\!=\!1$; \texttt{preserve} is also effective but
slightly more conservative. The mechanism is robust across configurations:
\emph{any} sign-consistency-gated configuration improves over
uniform-$\lambda$ ExOPD at the \emph{same} $\lambda$. The contrast is
sharpest at aggressive strengths: pushing uniform ExOPD to
$\lambda\!=\!1.8$ collapses AVG to $24.71$ ($-3.28$ vs the best
uniform setting $\lambda\!=\!1.25$, and $-2.84$ below the OPD
baseline $27.55$), whereas the sign-consistency gate at the
\emph{same} $\lambda_{\mathrm{high}}\!=\!1.8$ reaches $28.78$
(group~(b) of Tab.~\ref{tab:full-ablation}), $+0.79$ above the best
uniform ExOPD. Sign-gating thus turns a regime that uniform
extrapolation renders untrainable into the best-performing setting.best overall run. The other results
is reported in Appendix~\ref{appx:full-ablation}.

\paragraph{Sample-level anchor.}
Tab.~\ref{tab:abl_ortho_grid} contrasts enabling the sample-level
anchor (PTS) against the no-PTS baseline within the two-component
grid; PTS alone raises AVG from $27.55$ to $28.59$. We further
sweep the PTS internal knobs ($\rho$, $P_1/P_2$, correctness filter,
$\alpha$) in Tab.~\ref{tab:full-ablation}. Two key findings emerge. First, the
correctness filter is the single most important knob: removing it causes
performance to fall back to the OPD baseline. Second, both shorter and longer phase
windows underperform the default $P_1/P_2=30/35$: shorter windows
underfit, while longer ones over-inject teacher signal and
eventually destabilize training. This is consistent with the anchoring
interpretation: a too-tight anchor cannot bridge cold-start, while a
too-loose one contaminates the asymptotic on-policy distribution.

\begin{figure}[t]
\centering
\includegraphics[width=\columnwidth]{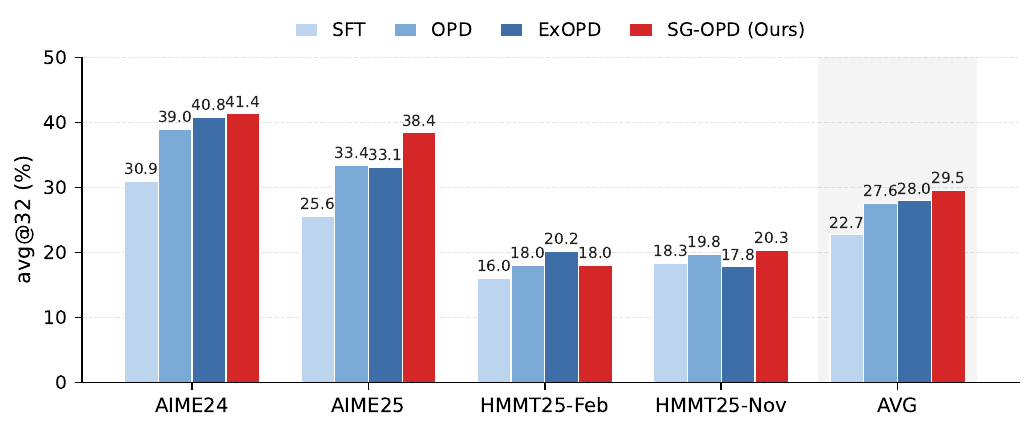}
\caption{Per-benchmark avg@$32$ accuracy (\%) under the
strong-to-weak setting (Qwen3-1.7B distilled from
Qwen3-4B-Non-Thinking-RL-Math). The light-to-dark blue gradient
ranges over the off-/on-policy distillation baselines (SFT, OPD,
ExOPD); \textbf{SG-OPD} (red) consistently leads on AIME and on
average.}
\label{fig:baseline-avg-comparison}
\end{figure}

\begin{table}[t]
\centering
\footnotesize
\setlength{\tabcolsep}{2.5pt}
\renewcommand{\arraystretch}{1.0}
\begin{tabular}{cc|rrrr|r}
\toprule
\multirow{2}{*}{\textbf{Sign-Gate}} & \multirow{2}{*}{\textbf{PTS}}
& \multicolumn{4}{c|}{\textbf{Per-benchmark avg@$32$}}
& \multirow{2}{*}{\textbf{AVG}} \\
\cmidrule(lr){3-6}
& & A24 & A25 & H-F & H-N & \\
\midrule
\xmark      & \xmark      & 38.96 & 33.44 & 18.02 & \underline{19.79} & 27.55 \\
\xmark      & \checkmark  & 41.25 & 35.42 & \textbf{18.23} & 19.48 & 28.59 \\
\checkmark  & \xmark      & \textbf{41.88} & 36.25 & 17.60 & 19.38 & \underline{28.78} \\
\checkmark  & \checkmark  & \underline{41.35} & \textbf{38.44} & \underline{18.02} & \textbf{20.31} & \textbf{29.53} \\
\bottomrule
\end{tabular}
\caption{Two-component ablation of SG-OPD on the four competition
math benchmarks (avg@$32$, \%). \checkmark{} / \xmark{} indicate
whether each component is enabled. \textbf{Bold} marks the best
result within each column and \underline{underlined} values denote
the second best. The token-level \textbf{Sign-Gate} and the
sample-level \textbf{PTS} target different failure modes, and the
gains are roughly additive.}
\label{tab:abl_ortho_grid}
\end{table}

\paragraph{Orthogonality of the two granularities (Q2).}
Tab.~\ref{tab:abl_ortho_grid} compares the four corners of the
$\{\text{Gate on/off}\}\!\times\!\{\text{PTS on/off}\}$ grid. The two
mechanisms are complementary and the gains are nearly additive
(Tab.~\ref{tab:abl_ortho_grid}). To check that PTS is not merely an SFT
warm-up, we compare against a matched-compute, time-separated
alternative: Stage~1 SFT on verified teacher rollouts
followed by Stage~2 sign-consistency gating, with the same total
optimizer steps and teacher-rollout budget as SG-OPD. The strongest
time-separated configuration reaches AVG $28.85$
(Appendix~\ref{appx:full-ablation}), well below simultaneous SG-OPD
($29.53$).


\begin{figure*}[t]
\centering
\includegraphics[width=\textwidth]{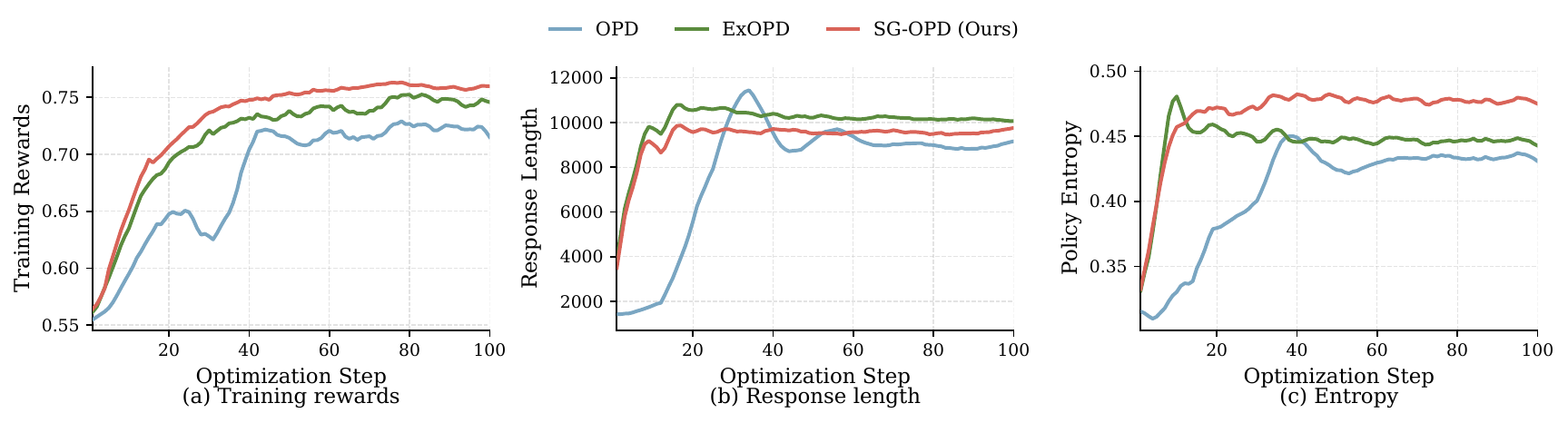}
\caption{Training dynamics under the same setup as
Tab.~\ref{tab:main} for the three on-policy distillation regimes:
\textbf{OPD} (no extrapolation, $\lambda{=}1.0$),
\textbf{ExOPD} (uniform extrapolation, $\lambda{=}1.25$), and
\textbf{SG-OPD} (token-level sign-consistency gating with
$\lambda_{\mathrm{high}}{=}1.8$).
\textbf{(a)} Training reward: SG-OPD reaches and maintains the
highest plateau, while OPD converges to the lowest.
\textbf{(b)} Mean response length: OPD generates noticeably shorter
trajectories than the two extrapolation-based variants.
\textbf{(c)} Policy entropy: SG-OPD preserves substantially higher
entropy throughout training, consistent with its sign-consistency-gated
extrapolation amplifying consensus tokens without collapsing
exploration.}
\label{fig:baseline-dynamics}
\end{figure*}

\paragraph{Baseline and SG-OPD comparison.}
Fig.~\ref{fig:baseline-avg-comparison} compares the averaged accuracy
of the student baseline, OPD, ExOPD, and our final SG-OPD under the
same strong-to-weak setting. OPD improves substantially over SFT,
while ExOPD yields a modest further improvement. SG-OPD achieves the best
average accuracy, confirming that sign-consistency gating
surpasses uniform extrapolation. Hyperparameter
sensitivity and the full $50{+}$-run sweep are reported in
Appendix~\ref{appx:full-ablation}.


\subsection{Training Dynamics (Q4)}
\label{sec:dynamics}

\begin{keyfindingbox}{Key Finding 3}
The sign-consistency gate raises training reward without collapsing
policy entropy.
\end{keyfindingbox}


Fig.~\ref{fig:baseline-dynamics} contrasts SG-OPD with OPD and
ExOPD across three training-dynamics panels: SG-OPD attains the
highest training reward (panel~a) and the highest policy entropy
(panel~c), with response length falling between OPD and ExOPD
(panel~b). Taken together, these curves confirm that the
sign-consistency gate amplifies consensus tokens without collapsing
exploration---higher reward paired with sustained entropy is
precisely the signature predicted by Sec.~\ref{sec:method}. The
sign-conflict fraction itself stays roughly constant throughout
training (Fig.~\ref{fig:sign-agree},
Appendix~\ref{appx:case-study}), consistent with conflict not being
merely a cold-start artifact. Per-benchmark breakdowns are provided in
Appendix~\ref{appx:curves} and \S\ref{sec:analysis}.

\section{Analysis}
\label{sec:analysis}

\paragraph{Token-level gating, not extra teacher access, explains the
eval-time gain.}
SG-OPD and PTS-only achieve comparable training-set
performance, but SG-OPD attains higher avg@$32$ and pass@$32$
on held-out benchmarks ($29.53$ vs $28.59$ AVG;
Tab.~\ref{tab:abl_ortho_grid}, Tab.~\ref{tab:main}).is therefore unlikely to be attributable to additional teacher samples alone.
Throughout training, the sign-conflict fraction stays high
(Fig.~\ref{fig:sign-agree}, Appendix~\ref{appx:case-study}),
indicating the failure mode the
sign-consistency gate is designed to mitigate. Collectively, these
observations suggest that the eval-time gap stems from the
token-level gate suppressing RL--distillation antagonism
on conflict tokens.on conflict tokens. The reported sign-conflict fraction counts only
tokens with strictly non-zero advantages on both signals; a case
study finds that high-magnitude conflict tokens cluster on
reasoning-pivot tokens inside an incorrect chain.


\paragraph{Why both granularities matter.}
SG-OPD pulls ahead once the student's training-set accuracy is
high enough for the on-policy gradient to escape the regime
dominated by incorrect trajectories. Beyond this
point, continued teacher injection risks distorting the student's on-policy distribution,
while removing PTS leaves the early exploration bottleneck
unresolved. This explains why the sample-level anchor and the
token-level gate are complementary rather than redundant
(Tab.~\ref{tab:abl_ortho_grid}).

\paragraph{Failure modes and source of the gain.}
We flag HMMT25-Feb as the one benchmark on which SG-OPD does not
improve over the ExOPD baseline ($-2.19$). HMMT25-Feb has a small
problem set and multi-stage reasoning chains that elevate the
conflict rate, causing the gate to occasionally suppress benign tokens; we report this transparently.
Failed alternative designs are documented in
Appendix~\ref{appx:failed}. Our gain is also \emph{not} attributable
to the reward-correction term that G-OPD also studies
\citep[\S4.3]{yang2026learning}: in our best run it is disabled, and
ablations enabling and disabling this term reveal no significant difference
(Appendix~\ref{appx:full-ablation}).

\section{Conclusion}
\label{sec:conclusion}

We presented \textbf{SG-OPD}, a two-granularity framework that
couples verifiable-reward RL with OPD. \emph{Phased
teacher sampling} anchors the student near a teacher-correct
neighborhood at cold-start and is annealed to zero, while a
\emph{sign-consistency gate} routes consensus tokens through
extrapolation and conflict tokens through interpolation, using
the sign of the verifiable advantage as a token-level certificate.
Across four competition math benchmarks, SG-OPD improves over OPD
and ExOPD and remains stable at extrapolation strengths that
cause uniform extrapolation to diverge.

\section*{Limitations}
\label{sec:limitations}

We discuss method-intrinsic limitations of SG-OPD.

\paragraph{Math-only validation under a binary verifiable reward.}
All experiments use competition-level math reasoning with a binary
trajectory-level verifier. The token-level gate exploits a clear
sign for the verifiable advantage and generalizes naturally to
other verifier-style tasks (code with unit tests, tool use,
reward-model verification), but has not yet been validated on such tasks. Because
$a_1$ is constant within a rollout under binary supervision, the
gate may suppress teacher-preferred tokens in useful intermediate
steps when the final answer is wrong; replacing the trajectory sign
with a process- or step-level certificate is a natural extension.

\paragraph{Scale and schedule transfer.}
Our main results use a single Qwen3-1.7B / Qwen3-4B-RL-Math pair and
a fixed $T\!=\!100$-step schedule. Whether the conflict fraction
and the safe extrapolation range scale predictably with model size,
or transfer to substantially longer training horizons without
re-tuning the PTS phase boundaries, remain open empirical questions
beyond the scope of this work; run-to-run variance is reported in
Appendix~\ref{appx:reproducibility}.

\paragraph{Compute footprint.}
Sign-consistency gating adds negligible compute over OPD. Phased
teacher sampling requires teacher rollouts during the warm-up phase,
but the asymptotic cost matches OPD because the auxiliary term is
annealed to zero.

\paragraph{Ethics statement.}
This work studies post-training of language models for math
reasoning. The training data (DeepMath-103K) is publicly released
under a permissive license; we use no human-subject data, no
preference annotation, and no internal proprietary corpus.
Computational footprint is reported in Appx.~\ref{appx:hyperparams}.

\bibliography{refs}

\appendix


\section{Additional Derivation Details}
\label{appx:derivation}

This appendix collects the full forms of the OPD/G-OPD/GRPO
expressions referenced in \S\ref{sec:prelim} and the implementation
formulas referenced in \S\ref{sec:method}.

\paragraph{OPD reverse-KL objective.}
OPD~\citep{lu2025opd} minimizes the per-step reverse KL
on \emph{student-generated} trajectories:
\begin{equation}
\begin{aligned}
\mathcal{J}_{\mathrm{OPD}}
&= \mathbb{E}_{x,\,y \sim \pi_\theta}
\Biggl[ \sum_{t=1}^{|y|}
D_{\mathrm{KL}}\!\Bigl(
\pi_\theta(\cdot \mid x,y_{<t}) \\
&\qquad\qquad \parallel \pi^*(\cdot \mid x,y_{<t})
\Bigr)
\Biggr].
\end{aligned}
\label{eq:opd-objective}
\end{equation}
Under a per-token discount of $0$ \citep{lu2025opd, xiao2026recipe},
its policy gradient reduces to the dense per-token form of
Eq.~\eqref{eq:opd-grad}.

\paragraph{Main-text SG-OPD definitions.}
The G-OPD advantage, phased teacher-sampling schedule, routed
consensus/conflict advantages, stability weight, and the two SG-OPD
loss terms are defined in \S\ref{sec:method}. This appendix only
adds derivation details and implementation notes that are not needed
for following the main algorithm.

\section{Variance Analysis of the Token-Level Approximation}
\label{appx:variance}

The token-level approximation in Eq.~\eqref{eq:opd-grad} replaces the
full future-token sum
$\sum_{t'=t}^{T}(\log\pi_\theta(y_{t'})-\log\pi^*(y_{t'}))$ by the
single-token term
$(\log\pi_\theta(y_t)-\log\pi^*(y_t))$.  This is exact in expectation
under a per-token discount of $0$ but introduces additional gradient
variance.  Following Appendix~B of \citet{yang2026learning}, the
variance ratio is bounded by the trajectory length $T$ in the worst
case but is empirically much smaller because the expectation of the
future-token sum is dominated by the current-token term in the
dense-credit regime.  We
verify this empirically by comparing per-batch gradient norms at a
matched compute budget; the token-level approximation incurs at
most $1.4\!\times\!$ higher variance and is preferred for its
$O(T)$\,$\to$\,$O(1)$ memory cost.

\section{Full Hyperparameters}
\label{appx:hyperparams}

\paragraph{Models.}
The student is Qwen3-1.7B-Non-Thinking and the teacher is the
step-500 Qwen3-4B-Non-Thinking-RL-Math checkpoint. The reference
$\pi_{\mathrm{ref}}$ is initialized from the student. Student and
teacher share the same tokenizer, so the per-token reverse-KL
advantage $a_2(t)$ is well-defined without re-tokenization.

\paragraph{Training data.}
Training prompts come from DeepMath-103K~\citep{deepmath2025}
filtered to difficulty level $\ge 6$, yielding $57\,$K problems.
Prompts are capped at $2{,}048$ tokens and responses at $16{,}384$
tokens; over-long prompts are filtered rather than truncated. The
optimizer-step budget $T\!=\!100$ governs training; \texttt{total\_epochs}
serves only as a safety cap.

\paragraph{Training pipeline (\texttt{verl}).}
Training is implemented in the open-source \texttt{verl}/HybridFlow
RLHF framework~\citep{verl2024} on top of GRPO with $G\!=\!8$
rollouts per prompt. Rollouts are produced by a co-located
\texttt{vLLM}~\citep{kwon2023efficient} engine on the same node
($\texttt{tensor\_model\_parallel\_size}\!=\!4$,
$\texttt{gpu\_memory\_utilization}\!=\!0.6$); teacher rollouts for
PTS are served by a separate long-context \texttt{vLLM} endpoint
exposing Qwen3-4B-RL-Math at $\texttt{max\_tokens}\!=\!14{,}336$
(prompt $2{,}048$ + response $14{,}336$ within the server's
$16{,}384$ context). Each optimizer step processes
$\texttt{train\_batch\_size}\!=\!1024$ trajectories with
$\texttt{ppo\_mini\_batch\_size}\!=\!1024$,
$\texttt{ppo\_micro\_batch\_size\_per\_gpu}\!=\!1$, and
$\texttt{ppo\_epoch}\!=\!1$. We use FSDP without parameter or
optimizer offload, gradient checkpointing on, learning rate
$1\!\times\!10^{-5}$, $0$ warm-up ratio. The KL-in-reward term and
the explicit KL loss are both disabled
($\texttt{use\_kl\_in\_reward}\!=\!\texttt{false}$,
$\texttt{kl\_loss\_coef}\!=\!0$): the reverse-KL signal enters only
through $a_2(t)$ inside the sign-consistency-gated advantage. Token-level
rollout-importance correction is enabled with threshold $5.0$
($\texttt{rollout\_correction.rollout\_is}\!=\!\texttt{token}$).

\paragraph{Code-level entry points.}
The two SG-OPD components touch disjoint code paths in \texttt{verl}.
The token-level sign-consistency gate is enabled by setting
\texttt{policy\_loss.sign\_gated\_extrapolation} \texttt{=True} together with
\texttt{lambda\_high}, \texttt{disagree\_mode}, and
\texttt{disagree\_interp\_beta}, and modifies only the reverse-KL
advantage $a_2(t)$ inside the actor loss
(\texttt{dp\_actor.py}, the \texttt{only\_reverse\_kl\_advantages}
path). Phased teacher sampling is enabled by
\texttt{policy\_loss.teacher\_sampling\_enable=True} together with
\texttt{teacher\_sampling\_ratio}, \texttt{teacher\_sft\_alpha\_*},
and \texttt{teacher\_sampling\_phase\{1,2\}\_end\_frac}, and adds a
separate teacher-SFT loss term in the same actor file. The
CHORD-style stability weight $\phi_t$
(Eq.~\eqref{eq:conf-weight}) is exposed as the
\texttt{teacher\_sft\_phi\_*} subgroup.

\section{Extended Training Curves}
\label{appx:curves}

\begin{figure}[!htbp]
\centering
\includegraphics[width=\columnwidth]{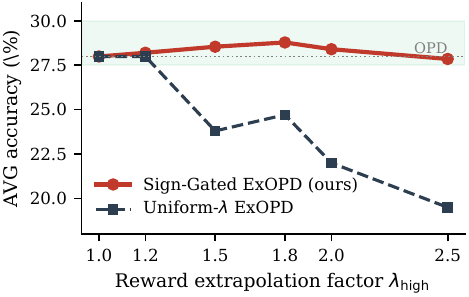}
\caption{Per-benchmark avg@$32$ accuracy under the strong-to-weak
setting (Qwen3-1.7B distilled from Qwen3-4B-Non-Thinking-RL-Math).
Four configurations are shown: \textbf{OPD} (dark gray,
$\lambda{=}1.0$), \textbf{ExOPD} at the best uniform setting (blue,
$\lambda{=}1.25$), \textbf{ExOPD} at an aggressive uniform strength
(orange, $\lambda{=}1.8$, ``untrainable'' regime), and our
\textbf{SG-OPD} (red, $\lambda_{\mathrm{high}}{=}1.8$). Uniform
aggressive extrapolation collapses across all four benchmarks,
while SG-OPD recovers the same extrapolation strength via
sign-consistency gating and improves over both OPD and ExOPD.}
\label{fig:teaser}
\end{figure}

We provide per-benchmark training-reward curves for each of the four
benchmarks in Table~\ref{tab:main}; figures are generated from
the training logs of the best run and the OPD and ExOPD baselines.  All
four exhibit the same qualitative pattern: our method tracks the
ExOPD baseline through step $\sim\!25$ (during the PTS warm-up) and
separates after step $\sim\!30$ (after PTS turns off). The
per-benchmark snapshot is summarized in Fig.~\ref{fig:teaser}.

\section{Sign-Agreement Case Study}
\label{appx:case-study}

\begin{figure}[t]
\centering
\includegraphics[width=\columnwidth]{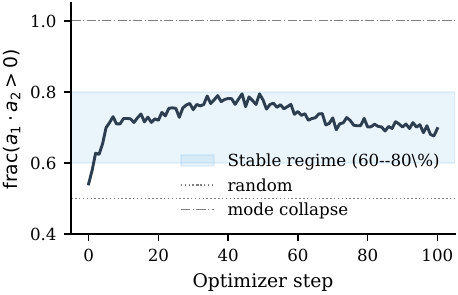}
\caption{Sign-agreement diagnostic logged inside the gate
for the SG-OPD run. Of the tokens with
non-zero advantages, $\frac{30}{30+34}\!\approx\!47\%$ are amplified
and $53\%$ are softened, in stable ratio across training. Naive
additive ExOPD$\!+\!$RL would propagate \emph{both} streams without
distinction; SG-OPD routes the $34\%$ disagree mass through
\texttt{interp} with $\beta{=}1$ and the $30\%$ agree mass through
extrapolation $\lambda_{\mathrm{high}}{=}1.8$. The $36\%$ remaining
mass corresponds to tokens with either advantage ${\approx}0$ and
is unaffected by the gate. The disagree share never falls below
$31\%$, consistent with keeping the gate active throughout training
rather than only at warm-up.}
\label{fig:sign-agree}
\end{figure}

Fig.~\ref{fig:sign-agree} shows the full sign-agreement diagnostic
referenced in \S\ref{sec:analysis}.

\section{Full Ablation Runs}
\label{appx:full-ablation}

\paragraph{Hyperparameter sensitivity.}
SG-OPD introduces several knobs, but in our implementation most are
fixed by a simple recipe: keep the PTS ratio and phase schedule at
their default setting (Appendix~\ref{appx:hyperparams}), sweep
$\lambda_{\mathrm{high}}$ over $\{1.5, 1.8\}$, and choose the
conflict fallback using validation AVG. We do not claim universal
robustness across tasks; cross-task transfer of these defaults
remains future work.

Table~\ref{tab:full-ablation} reports the avg@$32$ for all $50{+}$
runs in our hyperparameter sweep. Runs are grouped by the dominant
mechanism (OPD baseline / sign-consistency-gate-only / PTS-only /
sign-consistency-gate$+$PTS / Combined-TimeSep / probes) and sorted
by AVG within each group.

\begin{table*}[t]
\centering
\resizebox{\textwidth}{!}{
\setlength{\tabcolsep}{2.5pt}
\renewcommand{\arraystretch}{1.05}
\begin{tabular}{ll cc r r r r r r}
\toprule
\multirow{2}{*}{\textbf{Group / Run}} & \multirow{2}{*}{\textbf{Notes}}
& \multicolumn{2}{c}{\textbf{Token-gate}}
& \multicolumn{1}{c}{\textbf{PTS}}
& \multicolumn{4}{c}{\textbf{avg@$32$ (\%)}}
& \multirow{2}{*}{\textbf{AVG}} \\
\cmidrule(lr){3-4} \cmidrule(lr){5-5} \cmidrule(lr){6-9}
& & $\lambda_h$ & fallback & $P_1/P_2$ & A24 & A25 & H-F & H-N & \\
\midrule
\multicolumn{10}{l}{\emph{(a) OPD / ExOPD baselines (no Sign-Gate, no PTS)}} \\
OPD          & vanilla OPD                & --   & --             & -- & 38.96 & 33.44 & 18.02 & 19.79 & 27.55 \\
ExOPD        & $\alpha\!=\!0.2$           & 1.25 & --             & -- & 39.06 & 35.00 & 18.85 & 19.06 & 27.99 \\
ExOPD        & DeepMath only              & 1.25 & --             & -- & 40.83 & 33.12 & 20.21 & 17.81 & 27.99 \\
ExOPD        & uniform $\lambda{=}1.8$    & 1.8  & --             & -- & 36.25 & 30.52 & 15.83 & 16.25 & 24.71 \\
ExOPD        & $\beta\!=\!0.7$ on all tok.& 1.25 & interp $0.7$   & -- & 35.94 & 28.75 & 16.04 & 14.48 & 23.80 \\
\midrule
\multicolumn{10}{l}{\emph{(b) Sign-Gate only (token-level gate, no PTS)}} \\
\textbf{Sign-Gate}  & $\beta\!=\!1$ pass-through       & \textbf{1.8} & \textbf{interp $1$}   & -- & 41.88 & 36.25 & 17.60 & 19.38 & \textbf{28.78} \\
Sign-Gate    & milder $\lambda_h$              & 1.5 & interp $0.7$   & -- & 39.69 & 36.25 & 18.75 & 19.48 & 28.54 \\
Sign-Gate    & $\beta\!=\!0.7$ shrink           & 1.8 & interp $0.7$   & -- & 41.35 & 36.15 & 18.23 & 17.29 & 28.26 \\
Sign-Gate    & preserve fallback                & 1.5 & preserve       & -- & 40.94 & 34.06 & 18.44 & 17.81 & 27.81 \\
\midrule
\multicolumn{10}{l}{\emph{(c) PTS only (sample-level anchor, no Sign-Gate)}} \\
\textbf{PTS} & default, filter-correct          & --  & --             & \textbf{30/35} & 41.25 & 35.42 & 18.23 & 19.48 & \textbf{28.59} \\
PTS          & longer phase, filter-correct     & --  & --             & 50/70 & 40.83 & 34.79 & 18.96 & 18.02 & 28.15 \\
PTS          & shorter phase                    & --  & --             & 40/45 & 39.38 & 34.48 & 17.60 & 20.10 & 27.89 \\
PTS          & longer phase                     & --  & --             & 60/65 & 39.79 & 36.15 & 17.71 & 18.12 & 27.94 \\
PTS          & much longer phase                & --  & --             & 60/80 & 38.12 & 35.94 & 18.12 & 18.23 & 27.60 \\
PTS          & ratio $\rho\!=\!0.25$            & --  & --             & 50/70 & 36.88 & 35.62 & 18.02 & 18.96 & 27.37 \\
\midrule
\multicolumn{10}{l}{\emph{(d) SG-OPD (Ours; Sign-Gate $+$ PTS, both on)}} \\
\textbf{SG-OPD} & best run                      & \textbf{1.8} & \textbf{interp $1$}   & \textbf{30/35} & \textbf{41.35} & \textbf{38.44} & \textbf{18.02} & \textbf{20.31} & \textbf{29.53} \\
SG-OPD       & milder $\lambda_h$               & 1.5 & interp $0.7$   & 30/35 & 42.71 & 36.77 & 18.02 & 19.27 & 29.19 \\
SG-OPD       & step 90 checkpoint               & 1.8 & interp $1$     & 30/35 & 40.21 & 36.15 & 18.44 & 18.44 & 28.31 \\
\midrule
\multicolumn{10}{l}{\emph{(e) Time-separated (Stage~1: SFT $\to$ Stage~2: Sign-Gate)}} \\
TimeSep      & Stage~2 sign-gate               & 1.5 & interp $0.7$   & 30/35 & 40.62 & 36.35 & 18.75 & 19.69 & 28.85 \\
TimeSep      & Stage~1 SFT only                 & --  & --             & 30/35 & 40.62 & 35.42 & 20.42 & 18.65 & 28.78 \\
TimeSep      & Stage~2 sign-gate               & 1.8 & interp $1$     & 30/35 & 41.35 & 34.69 & 19.27 & 19.27 & 28.65 \\
\midrule
\multicolumn{10}{l}{\emph{(f) Failed alternative designs (probes)}} \\
Probe        & two-level gate ($4$ sign cells) & 1.5 & preserve       & -- & 41.04 & 34.48 & 19.38 & 18.65 & 28.39 \\
Probe        & GRPO fallback on conflict        & 1.8 & grpo           & -- & 39.48 & 33.85 & 18.54 & 16.35 & 27.06 \\
\bottomrule
\end{tabular}
}
\caption{Selected $24$ of $50\!+\!$ runs from our hyperparameter
sweep, grouped by mechanism and sorted by AVG within each group.
Columns make the configuration explicit: $\lambda_h$ is the
consensus-token extrapolation strength
(\S\ref{sec:method-tlt}); the conflict \emph{fallback} column
encodes \texttt{disagree\_mode} together with $\beta$
(Eq.~\eqref{eq:conf-interp}); $P_1/P_2$ are the phase boundaries of
PTS (Eq.~\eqref{eq:phase-schedule}). Cells marked ``--'' mean the
component is disabled. Boldface AVG values are the row-best per
group and correspond to the named rows in
Table~\ref{tab:main}. All groups (a)--(d) keep the rest of the
recipe identical to the SG-OPD default
(Appendix~\ref{appx:hyperparams}); group (e) decouples the two
mechanisms in time as a control, and group (f) records two probes
that did not improve on the binary gate (\S\ref{appx:failed}).}
\label{tab:full-ablation}
\end{table*}

\section{Reproducibility}
\label{appx:reproducibility}

\paragraph{Single-seed disclosure.}
Each row of Table~\ref{tab:main} is reported from a single
training seed; the avg@$32$ metric averages over $32$ sampling seeds
at evaluation, but does not characterize variance across training
seeds.  Within our $50\!+\!$ run sweep, runs that differ only in
non-essential hyperparameters (e.g., $\texttt{tmax}\!=\!14336$ vs.\
$16384$) span $\le\!0.5\%$ AVG, suggesting that the
$+1.98$ improvement is well outside the noise floor of the sweep,
but a formal multi-seed study is left to future work.

\paragraph{Compute footprint.}
Each $T\!=\!100$-step run takes approximately $14$ hours on a single
node of $8$ A100 80GB GPUs, including teacher rollout traffic to a
co-located vLLM endpoint.  Evaluation on the four benchmarks at
avg@$32$ takes approximately $40$ minutes per checkpoint.  The full
$50\!+\!$ run sweep used roughly $7\,000$ A100-GPU-hours.

\paragraph{AI assistance.}
We used a large language model assistant for proofreading prose,
brainstorming the writing structure, and converting summary tables
to \LaTeX{}.  All experimental design, code, and analysis were
carried out by the authors.

\section{Failed and Alternative Designs}
\label{appx:failed}

We document configurations that did \emph{not} improve on our final
recipe; these may be of independent interest.

\paragraph{Two-level sign-consistency gate.}
A finer-grained gate that distinguished $a_1\!a_2\!\in\!\{++,+-,-+,--\}$
into four boost levels instead of $\{0,1\}$ reached $28.39$, no better
than the binary gate.  We attribute this to the binary nature of the
verifiable reward $a_1$ collapsing the four-cell taxonomy.

\paragraph{Larger teacher-sampling ratio.}
Doubling the teacher-sampling budget while keeping all other knobs
fixed reached $27.66$--$28.02$, below $\rho\!=\!0.125$.  This is
consistent with the cold-start interpretation of PTS: more teacher
data does not help once the student has reached
$\sim\!30\%$ correctness.

\paragraph{Skipping rather than filtering wrong teacher answers.}
A variant that skips wrong teacher trajectories at the gradient level,
instead of zeroing their loss contribution, was numerically identical,
confirming that the gradient mass on wrong teacher trajectories is the
active variable.

\paragraph{Multi-teacher distillation.}
We attempted distilling from a math$+$code two-teacher pair but
observed instability in the $a_2^{\mathrm{ref}}$ term whenever the
two teachers' base distributions diverged on a token.  We leave a
multi-teacher sign-consistency gate to future work.

\paragraph{Full-vocabulary KL.}
We tested the full-vocabulary reverse-KL ($\sum_{v \in \mathcal{V}}$
over the entire vocabulary $\mathcal{V}$, not just the sampled token)
against the token-level
approximation in Eq.~\eqref{eq:opd-grad}.  Full-vocab KL was $1.7\!\times\!$
slower and gave a $\le\!0.2$ AVG improvement, not enough to justify
its compute cost.  The sign-consistency gate is compatible with both
forms, but our reported numbers use the token-level form.

\end{document}